\documentclass[journal,twoside,web]{ieeecolor}
\usepackage{generic}
\usepackage{cite}
\usepackage{amsmath,amssymb,amsfonts}
\usepackage[ruled,vlined]{algorithm2e}
\usepackage{graphicx}
\usepackage{booktabs}
\usepackage{hyperref}
\hypersetup{hidelinks=true}
\usepackage{textcomp}

\usepackage{multirow}

\def\BibTeX{{\rm B\kern-.05em{\sc i\kern-.025em b}\kern-.08em
    T\kern-.1667em\lower.7ex\hbox{E}\kern-.125emX}}
\begin{document}
\title{PathoSyn: Imaging-Pathology MRI Synthesis via Disentangled Deviation Diffusion}
\author{Jian Wang$^\dagger$,  Sixing Rong,  Jiarui Xing, Yuling Xu and Weide Liu, \IEEEmembership{Member, IEEE}
\thanks{Manuscript received XX XX, 2025; revised XX XX, 2025.}
\thanks{Jian Wang was with the Department of Radiology, Boston Children’s Hospital, Harvard Medical School, Boston, MA 02115 USA. (e-mail: jianbljh@gmail.com).}
\thanks{Sixing Rong is with the College of Science, Northeastern University, Boston, MA, 02115, USA. (e-mail: rong.s@northeastern.edu).}
\thanks{Jiarui Xing is with the School of Medicine, Yale University, New Haven, CT 06510 USA. (e-mail: jiarui.xing@yale.edu)}
\thanks{Yuling Xu is with the Department of Cardiac Surgery, The Second Affiliated Hospital of Jiangxi Medical College, Nanchang University, Nanchang 330000, China.}
\thanks{Weide Liu is with the College of Computing and Data Science, Nanyang Technological University, Singapore 639798. (e-mail: weide001@e.ntu.edu.sg)}
}

\maketitle

\begin{abstract}
We present PathoSyn, a unified generative framework for Magnetic Resonance Imaging (MRI) image synthesis that reformulates imaging-pathology as a disentangled additive deviation on a stable anatomical manifold. Current generative models typically operate in the global pixel domain or rely on binary masks, these paradigms often suffer from feature entanglement, leading to corrupted anatomical substrates or structural discontinuities. PathoSyn addresses these limitations by decomposing the synthesis task into deterministic anatomical reconstruction and stochastic deviation modeling. Central to our framework is a Deviation-Space Diffusion Model designed to learn the conditional distribution of pathological residuals, thereby capturing localized intensity variations while preserving global structural integrity by construction. To ensure spatial coherence, the diffusion process is coupled with a seam-aware fusion strategy and an inference-time stabilization module, which collectively suppress boundary artifacts and produce high-fidelity internal lesion heterogeneity. PathoSyn provides a mathematically principled pipeline for generating high-fidelity patient-specific synthetic datasets, facilitating the development of robust diagnostic algorithms in low-data regimes. By allowing interpretable counterfactual disease progression modeling, the framework supports precision intervention planning and provides a controlled environment for benchmarking clinical decision-support systems. Quantitative and qualitative evaluations on tumor imaging benchmarks demonstrate that PathoSyn significantly outperforms holistic diffusion and mask-conditioned baselines in both perceptual realism and anatomical fidelity. The source code of this work will be made publicly available.
\end{abstract}

\section{Introduction}
Pathological image generation is increasingly vital for medical image analysis, supporting data augmentation~\cite{frid2018synthetic,kebaili2023deep,koetzier2024generating}, disease progression modeling~\cite{yoon2023sadm,puglisi2024enhancing,ravi2022degenerative}, and clinical decision support~\cite{pinto2023artificial,comaniciu2016shaping}. However, acquiring balanced datasets is hindered by limited patient availability, ethical constraints, and high annotation costs. Generative models mitigate this scarcity by synthesizing realistic pathological images that complement real-world data. Beyond algorithmic utility, high-fidelity synthetic pathology aids in clinical intervention planning by illustrating plausible disease trajectories, surfacing rare edge cases, and enabling controlled evaluation of models prior to deployment~\cite{kim2022diffusion,shang2022learning,van2024synthetic,yoon2023sadm,puglisi2024enhancing}. Enhancing the fidelity and consistency of these images is therefore directly linked to developing safer, more reliable tools for clinical decision-making.

Despite growing interest, generating pathological images remains challenging due to the intrinsic asymmetry between anatomy and disease. The anatomical structure, including organ geometry and spatial layout, is largely stable for a given subject, whereas the pathological appearance is the main source of uncertainty, varying in intensity, texture, and clinical stage. The most clinically relevant variability is concentrated in the way disease perturbs a stable anatomical substrate. Existing generative models~\cite{koetzier2024generating,kazerouni2022diffusion} rarely make this distinction explicit. When operating directly in image space, anatomy and pathology are treated equally stochastic; the model must learn a high-dimensional distribution over all pixels simultaneously. This unnecessarily enlarges the solution space and permits the generator to introduce anatomical modifications in order to fit the data, thereby producing systematic distortions of anatomically stable structures. Consequently, current approaches fall into two suboptimal extremes: either they model the image as a holistic entity with no explicit separation~\cite{pinaya2022brain}, or they enforce complete separation via masking, which identifies location, but overlooks internal lesion appearance and continuity with surrounding tissue.
\begin{figure*}[!t]
\centering
\includegraphics[width=.80\textwidth]{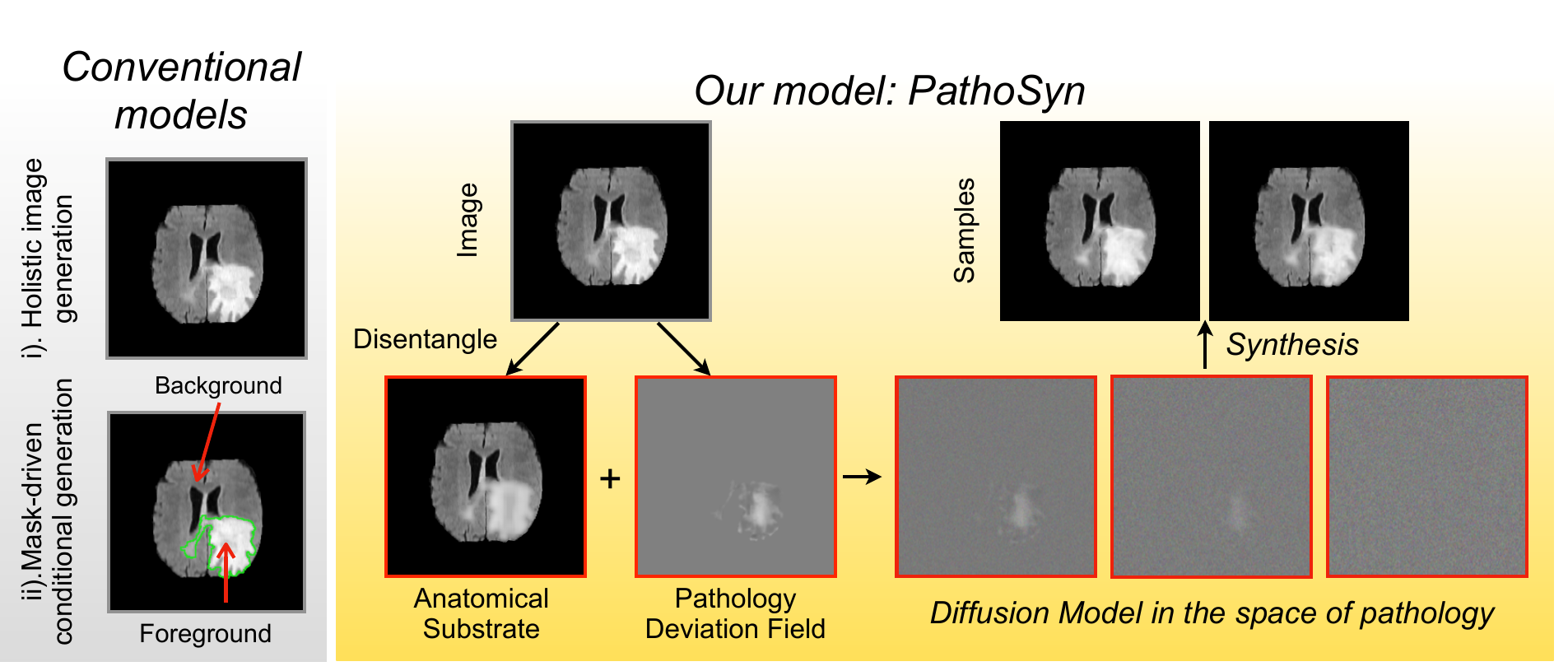}
     \caption{Conventional models synthesize images either by processing the entire image holistically or by fully segregating pathological from non-lesion regions. In contrast, PathoSyn explicitly disentangles anatomical structure from pathological alterations and performs generative modeling within a deviation space, thereby enabling controlled modulation of lesion characteristics while preserving the underlying anatomical substrate.}
\label{teaser}            
\end{figure*}

Although deep learning-based methods have significantly advanced visual fidelity compared to traditional rule-based algorithms, they have remained largely confined to the same image-space paradigm~\cite{jog2017random,ye2013modality}. Early applications of generative adversarial networks (GANs) synthesized lesions by modifying entire slices or volumes, sometimes conditioned on labels or coarse masks~\cite{goodfellow2020generative,isola2017image,wolterink2017deep,frid2018synthetic,yi2019generative}; these models can create sharp and visually plausible images, but tend to be unstable, difficult to control, and prone to hallucination of anatomy that does not correspond to any realistic background. Variational autoencoders (VAEs) introduced a probabilistic latent representation and more stable training~\cite{kingma2013auto,baur2021autoencoders,ehrhardt2022autoencoders}, and structured or disentangled variants attempted to separate anatomy and pathology in latent space~\cite{chartsias2019disentangled,cetin2023attri}. However, because decoding still happens directly to full images, they often oversmooth fine pathological details and leak stochastic variation into non-lesional regions. More recently, diffusion models have become the dominant paradigm for high-fidelity synthesis in both natural and medical imaging~\cite{ho2020denoising,rombach2022high,pinaya2022brain,kazerouni2022diffusion,ahsan2025comprehensive}. By iteratively denoising from a simple prior, they capture rich, multimodal image statistics and offer strong sample diversity. However, most medical diffusion models still learn a distribution over entire images or over mask-conditioned images, without explicitly isolating where uncertainty should reside. Mask conditioning helps to specify where pathology should appear but not what it should look like internally~\cite{ren2019mask,zhang2024lefusion,meng2024multi}; lesions may have plausible boundaries, but lack realistic internal texture, heterogeneity, or progression patterns. As a result, even advanced models still entangle anatomical background with pathological variation and fail to exploit that anatomy is largely deterministic while pathology is a structured deviation (Fig.~\ref{teaser}).

These limitations indicate that progress in pathological synthesis is not primarily determined by the choice of network architecture but by how the problem itself is represented. When a model treats the entire image as an unconstrained random variable, it must learn a full high-dimensional distribution in which anatomy and pathology vary simultaneously, even though only one of them is expected to change. Instead, a more principled formulation views a pathological image as the combination of two qualitatively different factors: a stable anatomical substrate that should remain preserved and a pathological deviation that carries the uncertainty, diversity and progression of the disease. Structurally separating these components reduces the effective complexity of the generative task and better reflects clinical reality: anatomy behaves as the baseline, while pathology introduces variation. In other words, disease does not replace the entire image; it perturbs an existing anatomical state.
Motivated by this perspective, we introduce PathoSyn, a deviation-based framework for pathological image generation. Rather than end-to-end synthesizing full images, PathoSyn focuses the generative process on the pathology component while preserving the underlying anatomical substrate. We instantiate this concept using a diffusion process defined over the deviation rather than the image itself, enabling pathological variation to be modeled as a controlled and spatially coherent modification that can be integrated with anatomy without introducing unintended structural change. This localizes stochasticity to where it belongs, within pathology, while maintaining anatomical fidelity by construction. As a result, PathoSyn captures clinically significant lesion heterogeneity, preserves global structure, and supports controllable manipulation of pathology appearance without destabilizing non-lesional regions.
In summary, PathoSyn provides three key contributions:
\begin{itemize}
    \item A representation shift that models pathology as a structured deviation from a preserved anatomical substrate, rather than treating the entire image as stochastic.
    \item A joint deep learning framework with deviation-space diffusion formulation that localizes generative uncertainty to disease regions, enabling controlled, interpretable, and anatomically consistent synthesis.
    \item A practical synthesis pipeline that produces realistic pathological variation while reducing anatomical distortion and improving the reliability of downstream analysis.
\end{itemize}
\section{Related Work}
Prior approaches to pathological image generation can be differentiated by the domain in which the generative distribution is defined and, consequently, which components of the image are treated as stochastic. 

\paragraph{Holistic Image Synthesis}
Early GAN- and VAE-based models~\cite{goodfellow2020generative,isola2017image,wolterink2017deep,frid2018synthetic,yi2019generative,kingma2013auto,baur2021autoencoders,ehrhardt2022autoencoders} assume a holistic generative process $\mathbf{x} \sim p_{\theta}(\mathbf{x})$, which requires the network to assign the probability mass across the entire image manifold $\mathcal{X} \in \mathbb{R}^{H \times W \times C}$. This implicitly treats both the anatomical substrate $\mathbf{a}$ and the pathological appearance $\mathbf{d}$ as coupled random variables. Because $p_{\theta}(\mathbf{x})$ must account for the morphology of the lesion without explicit structural constraints, the model often exploits anatomical degrees of freedom to minimize the training objective, leading to realistic pathology at the cost of distorted background anatomy.

\paragraph{Mask-Conditioned and Inpainting Methods}
To constrain the generative search space, mask-conditioned methods~\cite{ren2019mask,zhang2024lefusion,meng2024multi} narrow the problem to $\mathbf{x} \sim p_{\theta}(\mathbf{x} \mid \mathbf{m})$, where $\mathbf{m} \in \{0, 1\}^{H \times W}$ defines a binary lesion region. Although this specifies \textit{where} changes occur, it does not define \textit{how} the disease should manifest. The mask $\mathbf{m}$ encodes spatial support but lacks internal semantic identity; consequently, the generator solves a boundary-respecting fill-in problem rather than a clinically grounded deviation. This produces plausible contours but insufficient internal heterogeneity, as $\mathbf{m}$ resolves spatial ambiguity but not distributional uncertainty of the state of the disease.

\paragraph{Factorized Representation Learning}
Representation-based approaches introduce latent factors~\cite{bone2020learning}, often decomposing the image into $(\mathbf{z}_{anat}, \mathbf{z}_{path})$ to separate structural and pathological attributes. This aims to model $p_{\theta}(\mathbf{x})$ through a factorized prior $p(\mathbf{z}_{anat})p(\mathbf{z}_{path})$. However, the transformation back to the pixel space, $\mathbf{x} = f_{\theta}(\mathbf{z}_{anat}, \mathbf{z}_{path})$, inevitably recouples these factors. In practice, stochastic variation intended for the pathology leaks into non-lesional regions. While the representation is disentangled in latent space, the \textit{generative mapping} is not, meaning anatomical preservation is not guaranteed by construction.

\paragraph{Diffusion-based Generative Models}
Modern diffusion models improve stability and multimodal sampling, yet they typically evolve the stochastic process over the image $\mathbf{x}$ itself. Even context-aware variants implement denoising transitions $\mathbf{x}_t \rightarrow \mathbf{x}_{t-1}$ across the entire spatial domain. This enforces neither anatomical invariance nor deviation locality; the model effectively re-synthesizes the entire image rather than expressing uncertainty exclusively where pathology resides. Consequently, diffusion improves local fidelity while retaining the core representational limitation of image-space modeling.

In summary, existing methods either (i) conflate anatomy and pathology in a single stochastic space or (ii) enforce rigid spatial partitioning without modeling pathology as a continuous, anatomy-dependent deviation. Both reflect the same underlying issue: they attempt to learn $p_{\theta}(\mathbf{x})$ directly, while a clinically grounded formulation should localize uncertainty to the disease-bearing component $\mathbf{d}$ while treating anatomy $\mathbf{a}$ as a stable substrate. This motivates the shift toward our proposed deviation-space representation.

\section{Methodology}

\subsection{Disentangled Representation of Pathological Images}
We formulate generative modeling of pathological images by explicitly disentangling the subject-specific anatomical structure from disease-induced appearance variability. This decomposition is based on the distinct statistical properties of these two factors: for a given subject, the anatomical structure is largely invariant, while pathological manifestations exhibit high stochasticity, internal heterogeneity, and spatial localization, even within identical anatomical contexts.

In this framework, a pathological image $\mathbf{x} \in \mathbb{R}^{H \times W}$ is represented as the additive superposition of a deterministic anatomical substrate $\mathbf{x}_{\mathrm{sub}}$ and a stochastic pathological deviation field $\mathbf{r}$:
\begin{equation}
\mathbf{x} = \mathbf{x}_{\mathrm{sub}} + \mathbf{r},
\label{eq:conceptual_decomposition}
\end{equation}
where $\mathbf{x}_{\mathrm{sub}}$ characterizes the stable anatomical configuration, including organ morphology and global tissue organization, and $\mathbf{r}$ represents localized intensity deviations induced by pathological processes.

This additive formulation contrasts fundamentally with deformation-based models, such as metamorphosis, which account for pathological changes through geometric transformations of an underlying template. By isolating disease-related variation within the appearance space, our formulation enables the independent modeling of pathological features without distorting the underlying anatomical geometry. Consequently, $\mathbf{x}_{\mathrm{sub}}$ is treated as a subject-specific baseline, while $\mathbf{r}$ is interpreted as a structured residual that captures the uncertainty and progression of the disease.

The decomposition in Eq.~\eqref{eq:conceptual_decomposition} naturally leads to a conditional probabilistic formulation. Since the anatomical substrate is assumed to be preserved for a given subject, we model the deviation field as a random variable conditioned on both the anatomy and the spatial support of the lesion:
\begin{equation}
\mathbf{r} \sim p(\mathbf{r} \mid \mathbf{x}_{\mathrm{sub}}, \mathbf{m}),
\label{eq:conditional_distribution}
\end{equation}
where $\mathbf{m} \in \{0,1\}^{H \times W}$ denotes a binary mask defining the spatial domain of the pathology. By conditioning the distribution on $\mathbf{m}$, we explicitly constrain stochastic variation to the lesion site, thereby preventing unintended alterations to healthy tissue and ensuring anatomical fidelity by construction.

\subsection{Anatomical Substrate and Deviation Field Decomposition}

\paragraph{Inference of Latent Components}
Under the additive model defined in Eq.~\eqref{eq:conceptual_decomposition}, we assume that an observed pathological image $\mathbf{x}$ admits a latent decomposition $\mathbf{x} = \mathbf{x}_{\mathrm{sub}} + \mathbf{r}$. Let $\Omega \subset \mathbb{Z}^2$ denote the discrete image lattice, where $\mathbf{x} \colon \Omega \to \mathbb{R}$ represents the intensity function. The spatial support of the pathology is defined by a binary mask $\mathbf{m} \colon \Omega \to \{0,1\}$, where $\mathbf{m}(p)=1$ indicates that the pixel $p \in \Omega$ resides within the lesional region. We denote the complementary non-lesional mask as $\bar{\mathbf{m}} := \mathbf{1} - \mathbf{m}$.

Recovering $(\mathbf{x}_{\mathrm{sub}}, \mathbf{r})$ from a single observation $\mathbf{x}$ is an ill-posed inverse problem. To achieve a well-posed formulation, we impose structural constraints consistent with clinical priors: (i) intensities in the non-lesional domain $\bar{\mathbf{m}}$ are dominated by subject-specific anatomy, and (ii) pathological appearance variations are strictly localized to the support $\mathbf{m}$. These assumptions motivate a constrained inference scheme in which $\mathbf{x}_{\mathrm{sub}}$ is estimated primarily from the context of healthy tissue and $\mathbf{r}$ is limited to the pathological region.

\paragraph{Anatomical Substrate Estimation}
We define the anatomical substrate $\mathbf{x}_{\mathrm{sub}} \colon \Omega \to \mathbb{R}$ as a pathologically-suppressed representation that preserves subject-specific morphology. To enforce structural consistency outside the lesion, we utilize the Hadamard product $\odot$. The constraint $\mathbf{x}_{\mathrm{sub}} \approx \mathbf{x}$ on $\bar{\mathbf{m}}$ is formalized by a masked fidelity term $\|\left(\mathbf{x}_{\mathrm{sub}} - \mathbf{x}\right) \odot \bar{\mathbf{m}}\|_2^2$, where $\|\mathbf{y}\|_2^2 := \sum_{p \in \Omega} \mathbf{y}(p)^2$.

Because the underlying anatomy within the lesional region $\mathbf{m}$ is not observed, we introduce a counterfactual healthy reference $\mathbf{x}_{\mathrm{ph}} := \operatorname{Inp}(\mathbf{x}, \mathbf{m})$, where $\operatorname{Inp}(\cdot)$ denotes a fixed inpainting operator that extrapolates the anatomical context from $\bar{\mathbf{m}}$ to $\mathbf{m}$. We estimate $\mathbf{x}_{\mathrm{sub}}$ by minimizing the variational objective:
\begin{align}
\mathcal{L}_{\mathrm{sub}}(\mathbf{x}_{\mathrm{sub}}; \mathbf{x}, \mathbf{m}) &= \lambda_{\mathrm{out}} \|\left(\mathbf{x}_{\mathrm{sub}} - \mathbf{x}\right) \odot \bar{\mathbf{m}}\|_2^2 \nonumber  \\& + \lambda_{\mathrm{in}} \|\left(\mathbf{x}_{\mathrm{sub}} - \mathbf{x}_{\mathrm{ph}}\right) \odot \mathbf{m}\|_2^2,
\label{eq:Lsub_rigorous}
\end{align}
where $\lambda_{\mathrm{out}}, \lambda_{\mathrm{in}} > 0$ are hyperparameters that balance extrinsic fidelity and intrinsic regularization. In practice, $\mathbf{x}_{\mathrm{sub}}$ is parameterized by a neural predictor $f_{\theta}$ optimized for this objective.

\paragraph{Pathological Deviation Field}
Given the estimated substrate $\mathbf{x}_{\mathrm{sub}}$, the subject-specific deviation field is defined as the residual supported by the lesion:
\begin{equation}
\mathbf{r}_0 := (\mathbf{x} - \mathbf{x}_{\mathrm{sub}}) \odot \mathbf{m}.
\label{eq:r0_def_rigorous}
\end{equation}
By construction, $\mathbf{r}_0(p) = 0$ for all $p \in \bar{\mathbf{m}}$, ensuring that stochasticity is restricted to support of the lesion. Unlike deformation-based models that characterize pathology via non-rigid warps, $\mathbf{r}_0$ captures additive intensity and texture deviations in the appearance space. To bound the dynamic range and ensure numerical stability for the subsequent diffusion process, we apply a point-wise saturation operator:
\begin{equation}
\mathbf{r}_0 \leftarrow \delta \tanh\left(\frac{\mathbf{r}_0}{\delta}\right),
\label{eq:r0_saturation_rigorous}
\end{equation}
where $\delta > 0$ defines the maximum admissible deviation magnitude.

\subsection{Conditional Diffusion for Deviation Field Modeling}
Building upon the decomposition defined in Eq.~\eqref{eq:conceptual_decomposition}, we model the conditional distribution $p(\mathbf{r} \mid \mathbf{x}_{\mathrm{sub}}, \mathbf{m})$ using a Denoising Diffusion Probabilistic Model (DDPM)~\cite{ho2020denoising}. The diffusion process is defined over the deviation field $\mathbf{r}$ rather than the global image manifold $\mathbf{x}$. This formulation ensures that the generative modeling capacity is exclusively allocated to the pathological appearance, while the anatomical substrate remains deterministic and invariant.

\paragraph{Forward Diffusion Process}
Let $\mathbf{r}_0$ denote the clean deviation field extracted via Eq.~\eqref{eq:r0_def_rigorous}. We define a Markovian forward process that iteratively adds Gaussian noise over $T$ steps according to a variance schedule $\{\beta_t\}_{t=1}^T$:
\begin{equation}
q(\mathbf{r}_t \mid \mathbf{r}_{t-1}) = \mathcal{N}\left(\sqrt{1-\beta_t}\, \mathbf{r}_{t-1}, \beta_t \mathbf{I}\right), \quad t = 1, \dots, T.
\end{equation}
The latent state at any timestep $t$ can be expressed in closed form as:
\begin{equation}
\mathbf{r}_t = \sqrt{\bar{\alpha}_t}\, \mathbf{r}_0 + \sqrt{1 - \bar{\alpha}_t}\, \boldsymbol{\epsilon}, \quad \boldsymbol{\epsilon} \sim \mathcal{N}(0, \mathbf{I}),
\label{eq:rt_closed_form}
\end{equation}
where $\alpha_t = 1 - \beta_t$ and $\bar{\alpha}_t = \prod_{s=1}^t \alpha_s$. To ensure that stochasticity remains confined to the pathological domain, we enforce a spatial support constraint at each step:
\begin{equation}
\mathbf{r}_t \leftarrow \mathbf{r}_t \odot \mathbf{m}, \quad \forall t \in \{0, \dots, T\}.
\label{eq:mask_constraint}
\end{equation}
This projection prevents the diffusion process from introducing spurious variances or intensity shifts in the non-lesional anatomical regions.

\paragraph{Reverse Denoising and Training Objective}
The reverse process is parameterized by a conditional noise predictor $\boldsymbol{\epsilon}_\theta$, which estimates the noise component added to the deviation field. The network is conditioned on the noisy deviation $\mathbf{r}_t$, the anatomical substrate $\mathbf{x}_{\mathrm{sub}}$, the spatial mask $\mathbf{m}$, and the diffusion time step $t$. We optimize the parameters $\theta$ via the mean-squared error (MSE) objective:
\begin{equation}
\mathcal{L}_{\mathrm{diff}} = \mathbb{E}_{\mathbf{r}_0, \boldsymbol{\epsilon}, t} \left[ \left\| \boldsymbol{\epsilon} - \boldsymbol{\epsilon}_\theta(\mathbf{r}_t, \mathbf{x}_{\mathrm{sub}}, \mathbf{m}, t) \right\|_2^2 \right],
\label{eq:ddpm_loss}
\end{equation}
where $t \sim \mathcal{U}(1, T)$ and $\mathbf{r}_t$ is sampled according to Eq.~\eqref{eq:rt_closed_form} subject to the constraint in Eq.~\eqref{eq:mask_constraint}.

\paragraph{Ancestral Sampling and Synthesis}
During inference, synthesis begins with sampled Gaussian noise restricted to the lesion support: $\mathbf{r}_T \sim \mathcal{N}(0, \mathbf{I})$ followed by $\mathbf{r}_T \leftarrow \mathbf{r}_T \odot \mathbf{m}$. We iteratively recover the deviation field by applying the reverse transition for $t = T, \dots, 1$:
\begin{equation}
\mathbf{r}_{t-1} = \frac{1}{\sqrt{\alpha_t}} \left( \mathbf{r}_t - \frac{1 - \alpha_t}{\sqrt{1 - \bar{\alpha}_t}} \boldsymbol{\epsilon}_\theta(\mathbf{r}_t, \mathbf{x}_{\mathrm{sub}}, \mathbf{m}, t) \right) + \sigma_t \mathbf{z},
\label{eq:reverse_update}
\end{equation}
where $\mathbf{z} \sim \mathcal{N}(0, \mathbf{I})$ for $t > 1$ and $\mathbf{z} = \mathbf{0}$ for $t = 1$. To preserve anatomical fidelity, we re-apply the spatial projection $\mathbf{r}_{t-1} \leftarrow \mathbf{r}_{t-1} \odot \mathbf{m}$ after each update. The final sample $\hat{\mathbf{r}} = \mathbf{r}_0$ constitutes a stochastic realization of the pathological deviation conditioned on the subject-specific anatomy.

\subsection{PathoSyn: Deep Learning in the Deviation Space}

\begin{figure*}[!t]
\centering
\includegraphics[width=.90\textwidth]{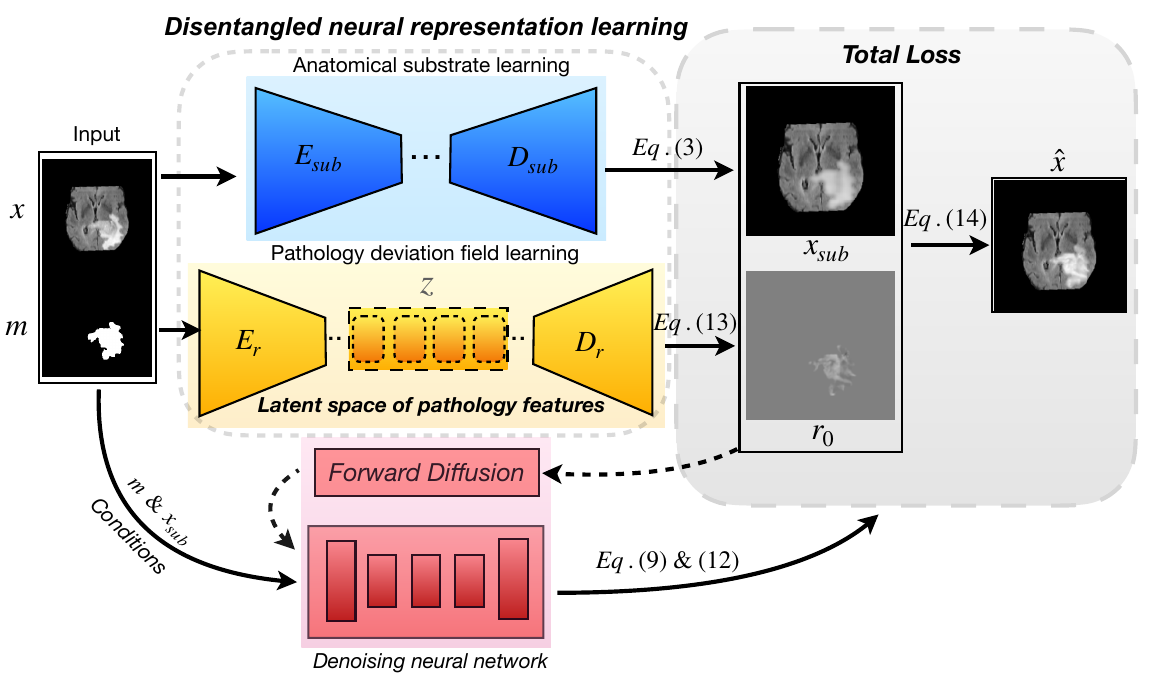}
     \caption{The PathoSyn architecture consists of two encoders and two decoders that disentangle an image into an anatomical substrate and a pathology deviation. The anatomical encoder–decoder pair learns a stable anatomical substrate that preserves underlying tissue geometry, while the pathology encoder–decoder pair models localized pathology deviation driven by lesion appearance. A generative diffusion module refines the pathology deviation to capture diverse and spatially coherent variations. A fusion layer recombines the anatomical substrate with the pathology deviation, enabling reconstruction and controllable synthesis of pathological images by adjusting lesion severity, extent, and spatial distribution while maintaining anatomical consistency. Please refer to the training and inference in Alg.~\ref{alg:train} and Alg.~\ref{alg:infer}.}
\label{netarch}            
\end{figure*}

We implement the proposed formulation through a unified deep learning framework that jointly estimates the anatomical substrate and models the conditional distribution of the pathological deviation field. All components are optimized under a shared objective, ensuring that anatomical-pathological decomposition and subsequent recomposition remain mutually consistent throughout training.

\paragraph{Network Architecture}
The anatomical substrate estimator $f_{\mathrm{sub}}$ is realized as a symmetric U-Net with four downsampling stages and skip connections, facilitating the preservation of high-frequency anatomical details while aggregating global contextual information. The diffusion noise predictor $\boldsymbol{\epsilon}_\theta$ utilizes a high-capacity U-Net architecture composed of Wide-ResNet blocks, with spatial self-attention integrated in the $16 \times 16$ bottleneck resolution to capture long-range dependencies within pathological regions. The diffusion time step $t$ is projected through sinusoidal positional embeddings and injected into each residual block to condition the denoising process.

\paragraph{Pathology-Deviation Regularization}
Although the diffusion objective in Eq.~\eqref{eq:ddpm_loss} learns the conditional distribution of pathological deviation fields, it does not explicitly enforce boundary smoothness or subject-specific consistency. We therefore introduce pathology-aware regularization to suppress spatial leakage and stabilize the interaction between the anatomical substrate and the deviation field.

Let $\mathbf{S} \in [0,1]^{H \times W}$ denote a soft blend map derived by applying a Gaussian kernel to the binary lesion mask $\mathbf{m}$, defining a narrow transition band in the periphery of the lesion. From $\mathbf{S}$, we derive a boundary weighting function:
\begin{equation}
\mathbf{w}_{\mathrm{ring}} = 4 \mathbf{S} (1 - \mathbf{S}),
\end{equation}
which emphasizes the lesion boundary while vanishing in both the interior and exterior domains. During training, we compute a denoised estimate of the clean deviation field:
\begin{equation}
\hat{\mathbf{r}}_0 = \frac{1}{\sqrt{\bar{\alpha}_t}} \left( \mathbf{r}_t - \sqrt{1 - \bar{\alpha}_t}\, \boldsymbol{\epsilon}_\theta(\mathbf{r}_t, \mathbf{x}_{\mathrm{sub}}, \mathbf{m}, t) \right).
\label{eq:r0_hat}
\end{equation}
We constrain $\hat{\mathbf{r}}_0$ to match the subject-specific deviation target $\mathbf{r}_0 = (\mathbf{x} - \mathbf{x}_{\mathrm{sub}}) \odot \mathbf{m}$ via a multi-term regularization loss:
\begin{align}
\mathcal{L}_{\mathrm{dev}} = &\ \lambda_{\mathrm{pat}} \left\| (\hat{\mathbf{r}}_0 - \mathbf{r}_0) \odot \mathbf{m} \right\|_1 \nonumber \\
&+ \lambda_{\mathrm{ring}} \left\| (\hat{\mathbf{r}}_0 - \mathbf{r}_0) \odot \mathbf{w}_{\mathrm{ring}} \right\|_1 \nonumber \\
&+ \lambda_{\mathrm{leak}} \left\| \hat{\mathbf{r}}_0 \odot (\mathbf{1} - \mathbf{m}) \right\|_1,
\label{eq:dev_loss}
\end{align}
where the terms, respectively, enforce internal fidelity, boundary consistency, and the suppression of pathological leakage into non-lesional regions.

\paragraph{Seam-Aware Recomposition}
To explicitly pair deviation modeling with image synthesis, a seam-aware recomposition layer is incorporated during training. Given the estimated anatomical substrate and deviation field, the synthesized pathological image is formed as follows:
\begin{equation}
\hat{\mathbf{x}} = \mathbf{x}_{\mathrm{sub}} + \mathbf{S} \odot \hat{\mathbf{r}}_0,
\label{eq:seam_synthesis_train}
\end{equation}
The consistency of reconstruction is enforced through $\mathcal{L}_{\mathrm{syn}} = \| (\hat{\mathbf{x}} - \mathbf{x}) \odot \mathbf{S} \|_1$, which prioritizes the supervision of anatomically sensitive transition zones to ensure seamless integration.

\paragraph{Overall Training Objective}
The final objective function jointly integrates anatomical reconstruction, diffusion modeling, and deviation regularization:
\begin{equation}
\mathcal{L} = \mathcal{L}_{\mathrm{sub}} + \lambda_{\mathrm{diff}} \mathcal{L}_{\mathrm{diff}} + \lambda_{\mathrm{dev}} \mathcal{L}_{\mathrm{dev}} + \lambda_{\mathrm{syn}} \mathcal{L}_{\mathrm{syn}}.
\label{eq:full_objective}
\end{equation}

\begin{algorithm}[t]
\small
\caption{PathoSyn Joint Training}
\label{alg:train}
\DontPrintSemicolon
\KwIn{Pairs $(\mathbf{x}, \mathbf{m})$; pre-trained $E_r, D_r$; steps $T$.}
\KwOut{Parameters for $f_{\mathrm{sub}}$ and $\boldsymbol{\epsilon}_\theta$.}

\While{not converged}{
    \tcp{1. Masking \& Substrate Extraction}
    $\bar{\mathbf{m}} \leftarrow \mathbf{1} - \mathbf{m}$; $\mathbf{S} \leftarrow \operatorname{Smooth}(\mathbf{m})$\;
    $\mathbf{x}_{\mathrm{sub}} \leftarrow f_{\mathrm{sub}}(\mathbf{x} \odot \bar{\mathbf{m}}, \bar{\mathbf{m}})$; $\mathbf{r}_0 \leftarrow (\mathbf{x} - \mathbf{x}_{\mathrm{sub}}) \odot \mathbf{m}$\;

    \tcp{2. Forward Diffusion }
    Sample $t \sim \mathcal{U}\{1, \dots, T\}$, $\boldsymbol{\epsilon} \sim \mathcal{N}(\mathbf{0}, \mathbf{I})$\;
    $\mathbf{r}_t \leftarrow (\sqrt{\bar{\alpha}_t} \mathbf{r}_0 + \sqrt{1 - \bar{\alpha}_t} \boldsymbol{\epsilon}) \odot \mathbf{m}$\;

    \tcp{3. Conditional Denoising \& Reconstruction}
    $\hat{\boldsymbol{\epsilon}} \leftarrow \boldsymbol{\epsilon}_\theta([\mathbf{r}_t, \mathbf{x}_{\mathrm{sub}}, \mathbf{m}], t)$ \tcp*{Condition on substrate}
    $\hat{\mathbf{r}}_0 \leftarrow \frac{1}{\sqrt{\bar{\alpha}_t}} (\mathbf{r}_t - \sqrt{1 - \bar{\alpha}_t} \hat{\boldsymbol{\epsilon}}) \odot \mathbf{m}$\;

    \tcp{4. Synthesis \& Loss Computation}
    $\hat{\mathbf{x}} \leftarrow \mathbf{x}_{\mathrm{sub}} + \mathbf{S} \odot \hat{\mathbf{r}}_0$\;
    $\mathcal{L}_{\mathrm{diff}} \leftarrow \|(\boldsymbol{\epsilon} - \hat{\boldsymbol{\epsilon}}) \odot \mathbf{m}\|_2^2$; $\mathcal{L}_{\mathrm{syn}} \leftarrow \|(\hat{\mathbf{x}} - \mathbf{x}) \odot \mathbf{S}\|_1$\;
    $\mathcal{L} \leftarrow \mathcal{L}_{\mathrm{sub}} + \lambda_{d} \mathcal{L}_{\mathrm{diff}} + \lambda_{v} \mathcal{L}_{\mathrm{dev}} + \lambda_{s} \mathcal{L}_{\mathrm{syn}}$\;
    
    Update $f_{\mathrm{sub}}$ and $\boldsymbol{\epsilon}_\theta$ via backpropagation\;
}
\end{algorithm}

\begin{algorithm}[t]
\small
\caption{PathoSyn Inference}
\label{alg:infer}
\DontPrintSemicolon
\KwIn{Image $\mathbf{x}$, mask $\mathbf{m}$; trained $f_{\mathrm{sub}}, \boldsymbol{\epsilon}_\theta$; steps $T$.}
\KwOut{Synthesized pathological image $\hat{\mathbf{x}}$.}

\tcp{1. Substrate Extraction \& Preparation}
$\bar{\mathbf{m}} \leftarrow \mathbf{1} - \mathbf{m}$; $\mathbf{S} \leftarrow \operatorname{Smooth}(\mathbf{m})$\;
$\mathbf{x}_{\mathrm{sub}} \leftarrow f_{\mathrm{sub}}(\mathbf{x} \odot \bar{\mathbf{m}}, \bar{\mathbf{m}})$\;

\tcp{2. Residual Noise Initialization}
Sample $\mathbf{r}_T \sim \mathcal{N}(\mathbf{0}, \mathbf{I})$; $\mathbf{r}_T \leftarrow \mathbf{r}_T \odot \mathbf{m}$ \tcp*{Initialize within mask}

\For{$t = T, \dots, 1$}{
    \tcp{Denoising conditioned on substrate and mask}
    $\hat{\boldsymbol{\epsilon}} \leftarrow \boldsymbol{\epsilon}_\theta([\mathbf{r}_t, \mathbf{x}_{\mathrm{sub}}, \mathbf{m}], t)$\;
    Sample $\boldsymbol{\epsilon}_{n} \sim \mathcal{N}(\mathbf{0}, \mathbf{I})$ if $t > 1$ else $\boldsymbol{\epsilon}_{n} \leftarrow \mathbf{0}$\;
    
    \tcp{Reverse Diffusion Step}
    $\mathbf{r}_{t-1} \leftarrow \frac{1}{\sqrt{\alpha_t}} \big( \mathbf{r}_t - \frac{1 - \alpha_t}{\sqrt{1 - \bar{\alpha}_t}} \hat{\boldsymbol{\epsilon}} \big) + \sigma_t \boldsymbol{\epsilon}_{n}$\;
    $\mathbf{r}_{t-1} \leftarrow \mathbf{r}_{t-1} \odot \mathbf{m}$ \tcp*{Enforce spatial support}
}

\tcp{3. Recomposition}
$\hat{\mathbf{x}} \leftarrow \mathbf{x}_{\mathrm{sub}} + \mathbf{S} \odot \mathbf{r}_0$ \tcp*{Combine substrate and residual}
\Return $\hat{\mathbf{x}}$\;
\end{algorithm}

\section{Experimental Evaluation}
\subsubsection{Dataset and Pre-processing}
Experiments are conducted on the BraTS 2020 brain tumor dataset~\cite{menze2014multimodal,bakas2017advancing,bakas2018identifying}, which comprises multi-institutional, multi-modal magnetic resonance images including T1, T1-weighted contrast-enhanced (T1Gd), T2-weighted and T2 Fluid Attenuated Inversion Recovery (FLAIR) volumes. The dataset provides expert voxel-level annotations for three tumor sub-regions: the necrotic and non-enhancing tumor core (NCR/NET), the peritumoral edema (ED), and the GD-enhancing tumor (ET). All imaging data are provided in a pre-processed format, including co-registration to a common anatomical template, interpolation to a uniform 1 $mm^3$ resolution, and skull-stripping.

We utilize the official training partition consisting of $n=369$ subjects. To ensure a rigorous evaluation framework and prevent data leakage, these subjects are divided at the patient level into three disjoint sets: a training set of $295$ subjects, a validation set of $37$ subjects and a hold-out testing set of $37$ subjects. While the training set is used to optimize the parameters of $f_{\mathrm{sub}}$ and $\boldsymbol{\epsilon}_\theta$, the validation set is used to monitor convergence and hyperparameter tuning. The final synthesis performance is evaluated exclusively on the testing set to ensure that the model is generalizable to unseen anatomical substrates. For computational efficiency, all 3D MRI volumes are processed into 2D axial slices and stored in HDF5 format, maintaining strict subject-level separation across all partitions.

All scans undergo standardized preprocessing, including skull stripping, spatial resampling to a common grid, and normalization of the intensity. To disentangle the effect of the quality of the synthesis from that of the size of the training set, all the augmentation regimes, comprising the baseline methods (Brain-LDM~\cite{pinaya2022brain}, MaskDiff-Inpaint~\cite{corneanu2024latentpaint}) are configured to contribute an equivalent number of synthetic samples to the training pool. To ensure a fair, controlled comparison and seamless integration with our proposed methods, we implement a pacth-wise VAE-GAN~\cite{gur2020hierarchical} within our unified PathoSyn framework. This allows us to standardize training protocols, architectural choices, and evaluation procedures across PathoSyn: VAE-GAN and our final model, PathoSyn: Diff, thus isolating the contribution of the generative paradigm rather than confounding factors from differing implementations.

\subsection{Implementation Details}
For methodological parity, all task-specific models share an identical 2D architecture and optimization pipeline. We employ the AdamW optimizer ($\eta = 10^{-4}$, weight decay $10^{-5}$) with a cosine annealing scheduler. Generative models are trained for 300 epochs, while downstream task models are trained for 200. PathoSyn-Diff utilizes a $T=1000$ step linear noise schedule with DDIM sampling during inference. PathoSyn-VAE-GAN employs a 128-dimensional latent space and a PatchGAN discriminator with an adversarial weight $\lambda_{\text{adv}} = 0.1$. Training is performed on NVIDIA RTX 3090/A6000 GPUs ($24\text{--}48$ GB VRAM). Inference utilizes test-time normalization only, without augmentation, to prevent masking potential domain shifts.

\subsection{Evaluation Framework}
\paragraph{Data Realism Analysis}
We quantify visual and statistical fidelity through a \textbf{real-versus-synthetic discrimination task}. A binary classifier is trained to distinguish authentic BRaTS 2023 slices from generated samples across all synthesis paradigms. Performance is measured via Receiver Operating Characteristic (ROC) curves and Area Under the Curve (AUC). 
\begin{itemize}
    \item 95\% bootstrap confidence intervals are used to evaluate statistical robustness.
    \item Lower AUC values ($\downarrow$) signify higher realism, indicating that synthetic samples are statistically closer to the real data manifold.
\end{itemize}

\paragraph{Mathematical Disentanglement Analysis}
To rigorously evaluate the separation of subject-specific anatomy and pathological variations, we analyze the statistical independence between the anatomical substrate $\mathbf{x}_{\text{sub}}$ and the deviation field $\mathbf{r}$. Ideally, a robust generative framework should achieve high-level feature orthogonality while maintaining low-level spatial alignment.
We define two metrics to quantify the disentanglement in the latent feature space $\Phi$:
\begin{enumerate}
    \item \textbf{Feature Orthogonality ($\mathcal{C}$):} We measure the absolute cosine similarity between the feature vectors $f_{\text{sub}} = \phi(\mathbf{x}_{\text{sub}})$ and $f_{\text{res}} = \phi(\mathbf{r})$ extracted by a frozen medical encoder $\phi(\cdot)$. A value $\mathcal{C} \approx 0$ indicates that the two representations reside in orthogonal subspaces.
    \item \textbf{Mutual Information (MI):} We estimate the non-linear dependency between $f_{\text{sub}}$ and $f_{\text{res}}$. A reduction in MI signifies that the stochastic pathological synthesis does not leak anatomical information into the deviation field.
\end{enumerate}
\paragraph{Downstream Task Utility}
We evaluate the functional utility of synthetic data by measuring performance gains in segmentation and classification tasks. All experiments are conducted in 2D to isolate the influence of generative quality from architectural advantages.
\begin{itemize}
    \item \textbf{Segmentation:} We benchmark across the supervised regimes (nnU-Net~\cite{isensee2021nnu}, U-Net~\cite{ronneberger2015u}), semi-supervised regimes (Mean-Teacher U-Net~\cite{tarvainen2017mean}), and unsupervised (Contra-Seg~\cite{liu2023contrastive}). The results are reported as mean Dice similarity coefficient (DSC) $\pm$ standard deviation.
    \item \textbf{Classification:} Models are assessed using Accuracy, AUC, and calibration error to determine whether synthetic augmentation improves generalization or introduces systematic bias.
\end{itemize}

\paragraph{Clinical Distributional Consistency}
To ensure that synthetic tumors respect clinical appearance statistics, we analyze feature-space alignment with real-world pathology:
\begin{itemize}
    \item \textbf{Perceptual Statistics:} Deep perceptual texture features are extracted to capture high-level semantic alignment.
    \item \textbf{Photometric reliability:} Intensity and texture distributions are assessed to verify mesoscopic consistency with real MRI sequences.
    \item \textbf{Statistical Proximity:} We analyze Empirical Cumulative Distribution Functions (CDFs) of feature distances; a shift toward the origin indicates a reduction in distributional shift from the authentic pathological manifold.
\end{itemize}

\subsection{Realism Evaluation via Discriminability and Stability}

\begin{figure}[hbt]
\centering
\includegraphics[width=.5\textwidth]{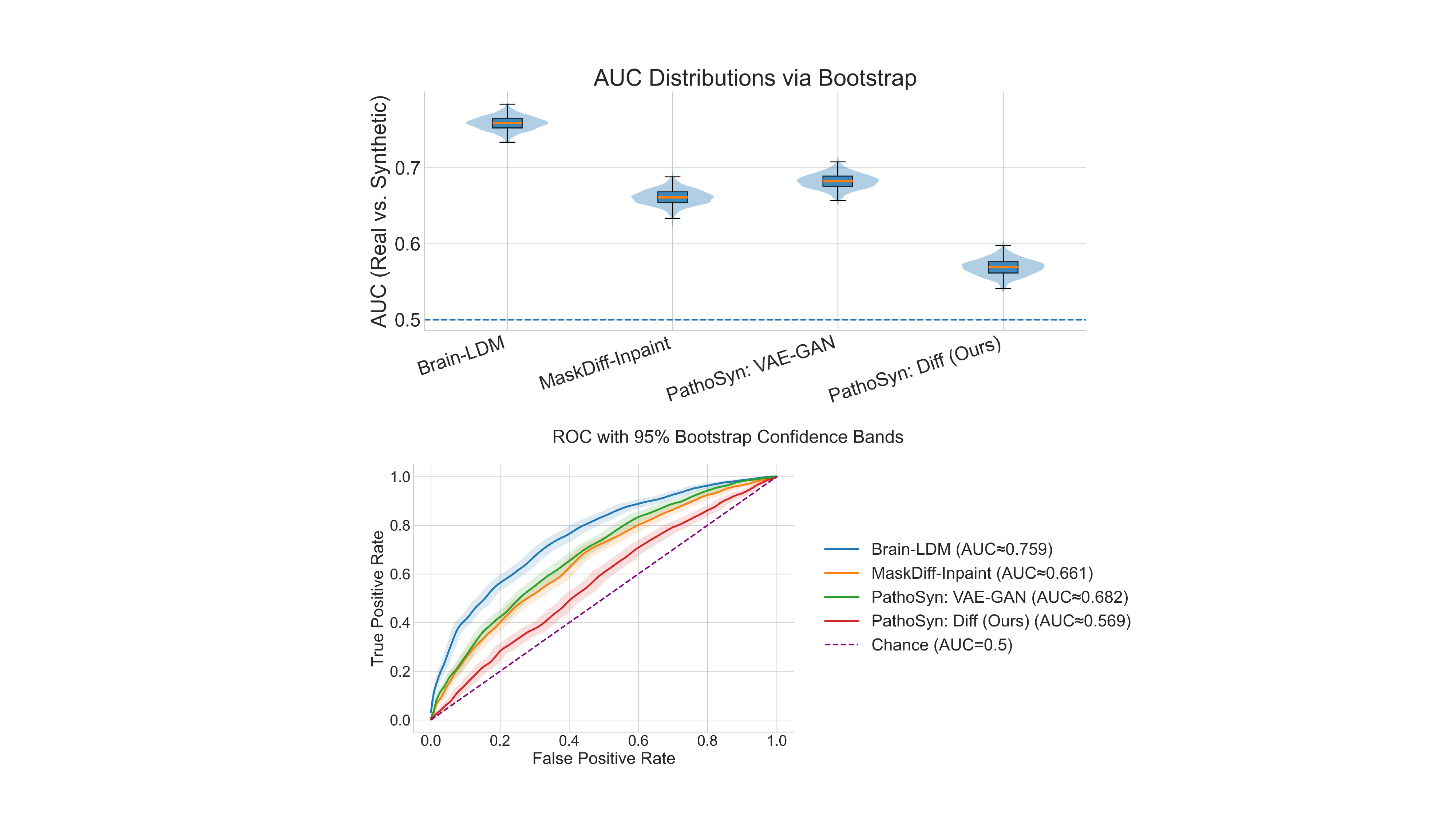}
     \caption{Top: ROC curves with 95\% bootstrap confidence intervals ($\downarrow$), where reduced detectability indicates greater alignment with real data. Bottom: Distribution of bootstrap AUC discriminability scores between real and generated images ($\downarrow$ is better).}
\label{AUC}            
\end{figure}

Fig.~\ref{AUC} (top) illustrates the performance of a binary discriminator trained to differentiate authentic BRaTS 2023 slices from synthetic samples. The results are presented as Receiver Operating Characteristic (ROC) curves, where shaded regions represent 95\% bootstrap confidence intervals. In this context, a lower Area Under the Curve (AUC $\downarrow$) indicates a reduced separability between the real and synthetic domains, signifying superior visual realism and minimized distributional bias. This convergence toward indistinguishability suggests that PathoSyn generates samples that reside closer to the authentic data manifold, thus mitigating the risk of introducing systematic biases during downstream clinical task training.

Fig.~\ref{AUC} (bottom) details the distribution of discriminability AUC values obtained via bootstrap resampling. The violin and box plots depict the variance in AUC estimates, while diamonds denote the point-estimate AUC derived from the mean ROC curve. Narrower distributions paired with lower medians ($\downarrow$) reflect improved generative stability and a consistent alignment with the real data distribution. This statistical stability across resamples demonstrates that the achieved realism is an inherent property of the model's learned representation rather than a stochastic artifact, ensuring robust generalization and minimal domain shift between synthetic and authentic pathological images.
\begin{figure*}[hbt]
\centering
\includegraphics[width=.88\textwidth]{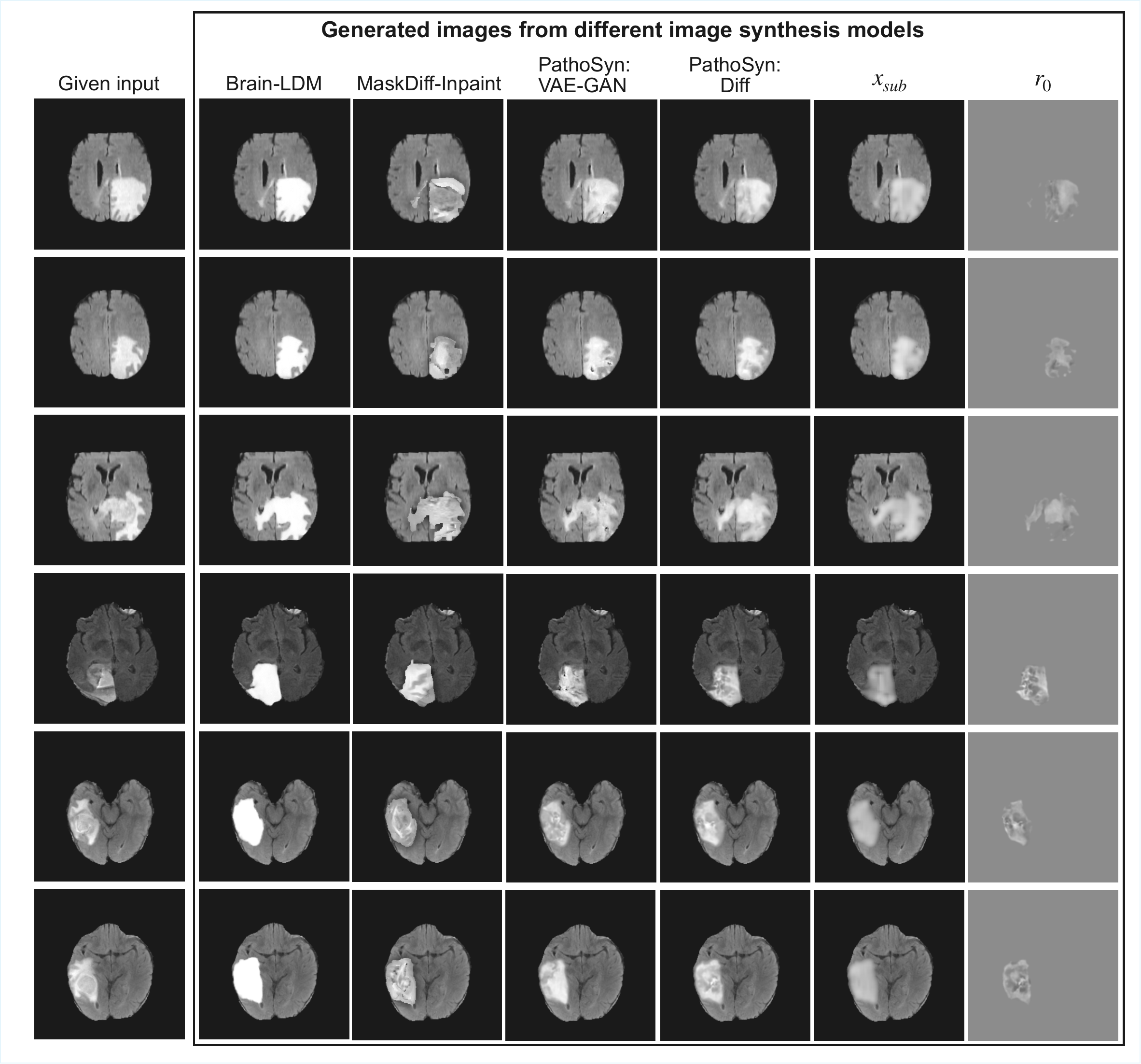}
     \caption{Qualitative comparison of pathological image synthesis. From left to right: the given image, Brain-LDM reconstruction, MaskDiff-Inpaint completion, PathoSyn (VAE-GAN), and PathoSyn (diffusion). The diffusion-based PathoSyn results further display the anatomical substrate and the corresponding pathology deviation fields, illustrating clearer separation between structural content and lesion-driven appearance, with improved spatial coherence and controllable pathology representation.}
\label{visual}            
\end{figure*}
\subsection{Mathematical Disentanglement Analysis}
As summarized in Tab.~\ref{tab:disentanglement}, we evaluate the mathematical independence between the anatomical substrate and pathological deviations.

\textbf{Anatomical Corruption in Holistic Models:} A fundamental flaw of holistic models like \textit{Brain-LDM} is the lack of explicit structural constraints. Since they treat the image as a monolithic distribution, the synthesis of a lesion inevitably leads to anatomical leakage, where the intensity and texture of healthy brain tissues are non-linearly corrupted. This is reflected in the high feature coupling (C=0.425), indicating that the model cannot alter the pathology without inadvertently modifying the identity of the underlying subject.

\textbf{Structured Disentanglement via Deviation Space:} Unlike dual-encoder frameworks (e.g., \textit{Meta-Auto}~\cite{bone2020learning}) that still exhibit significant latent overlap, PathoSyn achieves near-orthogonality (C=0.072). By constraining the generation to a dedicated deviation space, we effectively ``shield" the anatomical manifold from pathological stochasticity. This ensures that the digital twins generated are not only visually realistic, but also mathematically consistent, providing a safer and more reliable foundation for the training of clinical AI.
\begin{table}[t]
\centering
\caption{Quantitative Disentanglement Analysis. $\mathcal{C}$ and MI measure the independence between anatomy and pathology. Results are mean $\pm$ SD.}
\label{tab:disentanglement}
\resizebox{\columnwidth}{!}{
\begin{tabular}{lcccc}
\toprule
\textbf{Method} & \textbf{Mech.} & \textbf{Disent.} & \textbf{Cos Sim ($\mathcal{C}$)} $\downarrow$ & \textbf{MI} $\downarrow$ \\
\midrule
Brain-LDM \cite{pinaya2022brain} & Holistic & $\times$ & 0.425 $\pm$ .082 & 1.24 $\pm$ .15 \\
Meta-Auto \cite{bone2020learning} & Dual-Enc & \checkmark & 0.245 $\pm$ .041 & 0.78 $\pm$ .12 \\
MaskDiff-Inpaint & FG/BG Sep & \checkmark & 0.185 $\pm$ .038 & 0.65 $\pm$ .09 \\
\midrule
\textbf{PathoSyn (Ours)} & Dev-Space & \checkmark & \textbf{0.072 $\pm$ .018}$^{\dagger}$ & \textbf{0.21 $\pm$ .04}$^{\dagger}$ \\
\bottomrule
\end{tabular}}
\vspace{-2mm}
\end{table}

\subsection{Downstream Evaluation Protocol}
Tab.~\ref{tab:segmentation} summarizes Dice segmentation performance across four learning regimes, from fully supervised to unsupervised. Adding PathoSyn: Diff synthetic data yields consistent, statistically significant improvements ($p < 0.01$) for all architectures. On the supervised \textit{nnU-Net} baseline, PathoSyn: Diff reaches $0.845 \pm 0.021$ Dice, an absolute gain of $+6.5\%$ over ``No Augmentation.'' In the unsupervised \textit{Contra-Seg} regime, it improves performance by $+5.5\%$, indicating that PathoSyn effectively models the pathological manifold even without voxel-level labels. The lower standard deviation in PathoSyn cohorts further suggests that our disentangled deviation-space modeling provides more stable training signals than holistic methods like \textit{Brain-LDM}, which may introduce anatomical artifacts.
\begin{table*}[t]
\centering
\caption{Segmentation performance (Dice, $\uparrow$) for different synthesis strategies evaluated across four segmentation models spanning supervised (nnU-Net, U-Net), semi-supervised (Mean Teacher U-Net), and unsupervised (Contra-Seg) regimes. Results are reported as mean $\pm$ standard deviation. $^{\dagger}$ indicates statistically significant improvement over all non-PathoSyn baselines (paired $t$-test, $p < 0.01$).}
\label{tab:segmentation}
\addtolength{\tabcolsep}{4pt}
\scriptsize
\begin{tabular}{lcccc}
\toprule
\textbf{Synthesis Method} & 
\textbf{nnU-Net (Sup.)} & 
\textbf{U-Net (Sup.)} & 
\textbf{Mean Teacher (Semi-Sup.)} & 
\textbf{Contra-Seg (Unsup.)} \\ 
\midrule
No Augmentation            & 0.780 $\pm$ 0.035 & 0.770 $\pm$ 0.038 & 0.755 $\pm$ 0.045 & 0.740 $\pm$ 0.052 \\
Conventional Augmentation  & 0.800 $\pm$ 0.032 & 0.788 $\pm$ 0.035 & 0.770 $\pm$ 0.040 & 0.748 $\pm$ 0.049 \\
Brain-LDM                  & 0.810 $\pm$ 0.030 & 0.802 $\pm$ 0.032 & 0.785 $\pm$ 0.037 & 0.760 $\pm$ 0.045 \\
MaskDiff-Inpaint           & 0.815 $\pm$ 0.029 & 0.804 $\pm$ 0.031 & 0.790 $\pm$ 0.036 & 0.765 $\pm$ 0.043 \\
\midrule
\textbf{PathoSyn: VAE-GAN} & 0.830 $\pm$ 0.024$^{\dagger}$ & 0.820 $\pm$ 0.027$^{\dagger}$ & 0.805 $\pm$ 0.030$^{\dagger}$ & 0.780 $\pm$ 0.038$^{\dagger}$ \\
\textbf{PathoSyn: Diff}    & \textbf{0.845 $\pm$ 0.021}$^{\dagger}$ & \textbf{0.835 $\pm$ 0.025}$^{\dagger}$ & \textbf{0.820 $\pm$ 0.027}$^{\dagger}$ & \textbf{0.795 $\pm$ 0.035}$^{\dagger}$ \\
\bottomrule
\end{tabular}
\end{table*}

Tab.~\ref{tab:classification} summarizes the effect of different synthesis strategies on pathological state classification across three backbones: ResNet-50 (CNN), DenseNet (2D), and Swin Transformer. PathoSyn: Diff augmentation consistently yields the best performance, reaching a peak AUROC of 0.915 on ResNet-50, substantially outperforming holistic synthesis models such as \textit{Brain-LDM} and \textit{MaskDiff-Inpaint}. This highlights the benefit of modeling stochasticity within a constrained deviation space rather than across the full image manifold. PathoSyn: Diff also delivers the best calibration, with the lowest Expected Calibration Error (ECE) across all test regimes (0.043–0.045). While conventional and holistic augmentations increase AUROC, Tab.~\ref{tab:classification} shows they offer only modest ECE gains. In contrast, our diffusion-based approach reduces calibration error by about 35\% relative to No Augmentation. This indicates that high-fidelity, anatomically grounded lesions from PathoSyn prevent overconfidence on synthetic artifacts, yielding more reliable probability estimates crucial for clinical decision support.
\begin{table*}[t]
\centering
\caption{Classification performance under different synthesis strategies evaluated across CNN, Volumetric, and Transformer architectures. AUROC ($\uparrow$) measures discriminative capability, while Expected Calibration Error (ECE, $\downarrow$) assesses the reliability of probability estimates. Results are reported as mean $\pm$ standard deviation. $^{\dagger}$ denotes statistically significant improvement over all non-PathoSyn baselines (paired $t$-test, $p < 0.01$).}
\label{tab:classification}
\addtolength{\tabcolsep}{-1pt}
\scriptsize
\begin{tabular}{lcccccc}
\toprule
\multirow{2}{*}{\textbf{Synthesis Method}} & \multicolumn{2}{c}{\textbf{ResNet-50 (2D CNN)}} & \multicolumn{2}{c}{\textbf{DenseNet (2D)}} & \multicolumn{2}{c}{\textbf{Swin Transformer}} \\
\cmidrule(lr){2-3} \cmidrule(lr){4-5} \cmidrule(lr){6-7}
& AUROC $\uparrow$ & ECE $\downarrow$ & AUROC $\uparrow$ & ECE $\downarrow$ & AUROC $\uparrow$ & ECE $\downarrow$ \\
\midrule
No Augmentation            & 0.860 $\pm$ 0.022 & 0.070 $\pm$ 0.011 & 0.855 $\pm$ 0.024 & 0.068 $\pm$ 0.012 & 0.852 $\pm$ 0.023 & 0.066 $\pm$ 0.013 \\
Conventional Augmentation  & 0.880 $\pm$ 0.020 & 0.062 $\pm$ 0.010 & 0.875 $\pm$ 0.021 & 0.060 $\pm$ 0.011 & 0.870 $\pm$ 0.021 & 0.059 $\pm$ 0.011 \\
Brain-LDM                  & 0.885 $\pm$ 0.019 & 0.060 $\pm$ 0.010 & 0.880 $\pm$ 0.019 & 0.058 $\pm$ 0.010 & 0.875 $\pm$ 0.020 & 0.057 $\pm$ 0.010 \\
MaskDiff-Inpaint           & 0.890 $\pm$ 0.018 & 0.058 $\pm$ 0.009 & 0.885 $\pm$ 0.018 & 0.056 $\pm$ 0.009 & 0.880 $\pm$ 0.019 & 0.055 $\pm$ 0.009 \\
\midrule
\textbf{PathoSyn: VAE-GAN} & 0.905 $\pm$ 0.016$^{\dagger}$ & 0.050 $\pm$ 0.008$^{\dagger}$ & 0.900 $\pm$ 0.016$^{\dagger}$ & 0.048 $\pm$ 0.009$^{\dagger}$ & 0.895 $\pm$ 0.017$^{\dagger}$ & 0.047 $\pm$ 0.009$^{\dagger}$ \\
\textbf{PathoSyn: Diff}    & \textbf{0.915 $\pm$ 0.015}$^{\dagger}$ & \textbf{0.045 $\pm$ 0.008}$^{\dagger}$ & \textbf{0.910 $\pm$ 0.015}$^{\dagger}$ & \textbf{0.044 $\pm$ 0.008}$^{\dagger}$ & \textbf{0.905 $\pm$ 0.015}$^{\dagger}$ & \textbf{0.043 $\pm$ 0.008}$^{\dagger}$ \\
\bottomrule
\end{tabular}
\end{table*}

\begin{figure*}[hbt]
\centering
\includegraphics[width=.97\textwidth]{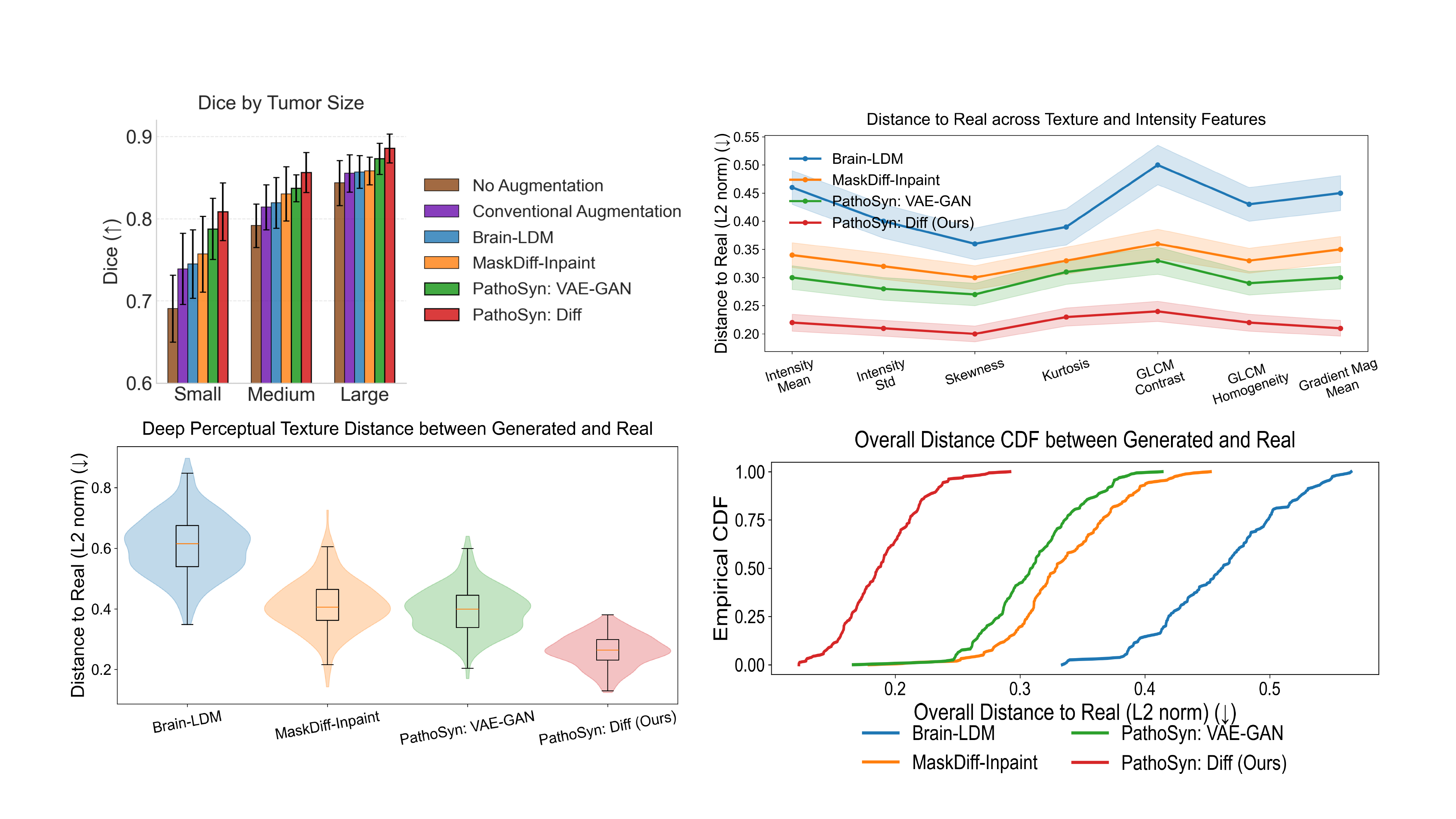}
     \caption{Top left: segmentation performance on generated images from multiple models, including no augmentation and conventional augmentation baselines, reported by Dice coefficients to evaluate downstream utility. Top right: deep perceptual texture statistics quantifying alignment between generated and real images in feature space. Bottom left: distributions of intensity- and texture-based features comparing generated versus real samples, reflecting low-level photometric and mesoscopic consistency. Bottom right: empirical cumulative distribution functions of feature distances, with the x-axis denoting deviation from real data; curves concentrated toward the origin indicate improved fidelity and reduced distributional shift.}
\label{comp}            
\end{figure*}

Fig.~\ref{comp} (top left) shows tumor segmentation performance for models trained with different synthesis strategies. Beyond overall gains, we assess sensitivity to tumor size. Most generative methods offer modest improvements for large, high-contrast lesions but degrade sharply for smaller ones, largely due to loss of high-frequency texture in the latent space. In contrast, PathoSyn: Diff maintains strong performance across all lesion sizes. For \textit{small tumors} (lowest quartile of lesion volume), it achieves the largest relative accuracy gain. This stems from its localized deviation-space modeling, which avoids the blurring seen in holistic models such as \textit{Brain-LDM}. By restricting diffusion to the pathological support $m$, PathoSyn preserves sharp boundaries and mesoscopic textures essential for detecting early-stage or small lesions. Together, Tab.~\ref{tab:segmentation} and Fig.~\ref{comp} show that by generating high-fidelity synthetic examples of typically underrepresented small lesions, PathoSyn mitigates voxel-level class imbalance and yields more calibrated, more sensitive tumor segmentation.

\subsection{Clinical Distributional Consistency}

Fig.~\ref{visual} qualitatively compares the clinical plausibility of generated lesions. A key requirement in neuro-oncology is respecting anatomical boundaries and tissue-specific intensity profiles. Holistic models like \textit{Brain-LDM} often generate ``low-level'' lesions that ignore underlying textures, whereas PathoSyn: Diff ensures that deviations—such as peritumoral edema—follow white matter tracts and ventricular constraints. Preserving the anatomical substrate during diffusion makes the synthesized pathology appear physiologically anchored rather than superimposed.

Fig.~\ref{comp} (top right) reports Intensity and Texture Consistency, including first-order statistics (Mean, Skewness, Kurtosis) and second-order GLCM metrics (Contrast, Homogeneity). High-grade gliomas are radiologically defined by pronounced heterogeneity, with high intensity variance and characteristic gradients at infiltrative margins. The close match between our results and clinical data across these descriptors shows that PathoSyn accurately models the lesion’s internal photometric structure. This textural fidelity is crucial for training diagnostic models that are sensitive to subtle tissue variations used to determine tumor grade.

Fig.~\ref{comp} (bottom left) measures the Deep Conceptual Texture Distance as a proxy for high-level semantic alignment. Radiologists characterize gliomas by complex morphologies, including internal mottling, rim enhancement, and central necrosis. Holistic models tend to yield over-smoothed or biologically sterile textures, but PathoSyn: Diff achieves the smallest conceptual distance to the BRaTS distribution, indicating that its latent space captures the semantic markers critical for diagnosis. This high-level alignment helps prevent shortcut learning: if synthetic data contain digital artifacts or lack biological complexity, downstream models may learn fake cues instead of real pathology. By narrowing this conceptual gap, PathoSyn encourages segmentation and classification networks to learn the same radiomic signatures found in patients, improving clinical reliability and generalization to real-world workflows.

Fig.~\ref{comp} (bottom right) shows the Empirical CDF of the global feature distance to the real pathological manifold, providing a population-level view of generative reliability. The leftward shift of the PathoSyn curve indicates that samples consistently fall within the clinically acceptable region of the true distribution. Clinically, this distributional stability means the framework does not just produce occasional high-quality images but reliably generates diverse, biologically credible digital twins, reducing the risk of outliers that could undermine downstream clinical AI robustness.

\section{Discussion \& Conclusion}
In this paper, we introduce PathoSyn, a pathology-aware medical image synthesis framework that disentangles stable anatomical structure from stochastic pathological variation. Instead of synthesizing complete images from a single joint distribution, PathoSyn represents disease as a deviation field superimposed on the preserved anatomy. This formulation limits identity leakage, protects the patient's specific structure, and improves the fidelity of tumor-related appearance changes. Compared to state-of-the-art approaches (holistic diffusion models, mask-based inpainting, and VAE–GANs), PathoSyn produces images with sharper infiltrative tumor margins, more realistic edema and necrotic textures, and fewer structural distortions. The low empirically measured correlation between anatomical and pathological representations (C = 0.072) indicates that pathology can be modulated with minimal unintended alteration of the anatomy of the core brain. Our framework removes the need for manual post hoc blending or heuristic inference-time adjustments, enabling seamless synthetic images for downstream research and clinical assessment. More broadly, the deviation field formulation provides a conceptual tool for modeling structured variation in computer vision and machine learning. It enables the representation of subtle changes in internal features and localized textural shifts without altering the global geometry. This principle extends to other medical imaging modalities such as CT, PET, ultrasound, and digital pathology, where disease progression often appears as regional texture and intensity changes rather than large-scale shape deformation.

A prominent clinical use case for PathoSyn is preoperative planning for brain tumor surgery. By producing multiple plausible MRI variants for an individual patient, PathoSyn can support clinicians in anticipating the short-term evolution of tumor texture, infiltration patterns, and peritumoral edema in the interval preceding surgery. These synthetic scenarios may help neurosurgeons and radiologists in the operative planning, risk stratification, and evaluation of alternative strategies in settings where updated images are unavailable or when rapid disease progression is suspected. However, the current implementation still has potential for further refinement. PathoSyn primarily models appearance changes and does not yet simulate large-scale structural deformations such as mass effect, ventricular compression, or midline shift. In addition, robustness across scanners, institutions, and acquisition protocols remains constrained by domain shift, indicating a need for improved domain adaptation and harmonization. These issues must be resolved before the framework can be considered for regulated clinical deployment.

Future work will extend PathoSyn into a unified generative framework capable of jointly modeling both appearance alterations and geometric deformations. This expansion will enable representation of not only tumor texture but also the spatial impact of neoplastic growth on adjacent brain structures. The overarching objective is to support realistic data generation for surgical planning, longitudinal disease monitoring, and robustness evaluation of machine learning models. By explicitly combining anatomical stability with controlled, realistic pathological variation, PathoSyn constitutes a step toward clinically meaningful synthetic data generation that can underpin more reliable, interpretable, and trustworthy medical AI systems.
\section*{Conflict of Interest}
The authors declare no conflicts of interest.

\section{Reference}
\bibliographystyle{IEEEtran}
\bibliography{reference}
\end{document}